# Automated machine vision control system for technological nodes assembly process

Nikolay Shtabel[a)], Mikhail Saramud[b)], Stepan Tkachev [c)] and Iakov Pikalov [d)]

*Reshetnev Siberian State University of Science and Technology, Krasnoyarsk, Russia*

[a)]shtabnik@gmail.com
[b)]Corresponding author: msaramud@mail.ru
[c)]steep_st@mail.ru
[d)]yapibest@gmail.com

**Abstract.** The paper discusses the prerequisites for the creation, technical solutions and implementation of an automated control system for the assembly of a small spacecraft. Both the hardware and software implementation of the system that provides control and logging of the assembly process of individual units at various workplaces are analyzed. The article presents solutions to reduce the requirements for equipment used to control the assembly technology, in particular, to use cameras with a lower resolution, through the use of special algorithms for the formation and processing of technological marks. A tool is presented that allows you to control the tightening torques of threaded connections and limit the tightening torque according to a given algorithm with wireless control. The developed system provides the functions of not only control, but also logging of the technological process, which can be useful in the future when creating a digital twin of the product.

## INTRODUCTION

Currently many assembly operations in small-scale and piece production of aerospace products are performed manually, since the use of automation for small volumes of production is not economically feasible [1]. The low degree of automation of production processes is one of the factors leading to the possible occurrence of errors in the assembly of individual units and components, disruption of the technological process.

The control of manual assembly operations is also often carried out by an employee, manually or using special stands. On the one hand, this method of control allows to detect errors in the technological process, on the other hand, it does not completely exclude the human factor, since control is often carried out selectively and can be carried out in violation of procedures, which will not be fixed in any way.

With a small assembly series of the order of several hundred products per year, as an alternative to full-fledged automation of the assembly, for example, using robotic manipulators, the possibility of partial automation is considered, which includes the use of specialized equipment and tools at each workplace. Also, to control the correctness of assembly operations, it is advisable to use machine vision cameras and specialized software for quality control of work.

This paper considers one of the options for an automated control system that allows you to control the correct execution of technological operations, automate individual operations, such as tightening screw connections. The advantage of the proposed system is the possibility of its implementation both at existing and newly created workplaces, which makes it possible to implement assembly control at all stages of manufacturing a particular product, without being tied to the automation schedule of the entire process.

As a prototype of an automated assembly line, the possibility of automating the assembly of small spacecraft is considered, which is relevant in connection with the spread of the concept of low-orbit space constellations of the same type of low-cost satellites for organizing terrestrial communications [2,3].

# ORGANIZATION OF AUTOMATED CONTROL OF THE ASSEMBLY OF SMALL SPACE VEHICLES

The authors developed and implemented an experimental system for automated assembly control at the workplaces of small-scale production of a small spacecraft. The scheme of the experimental site for assembling a small spacecraft is shown in Fig. 1.

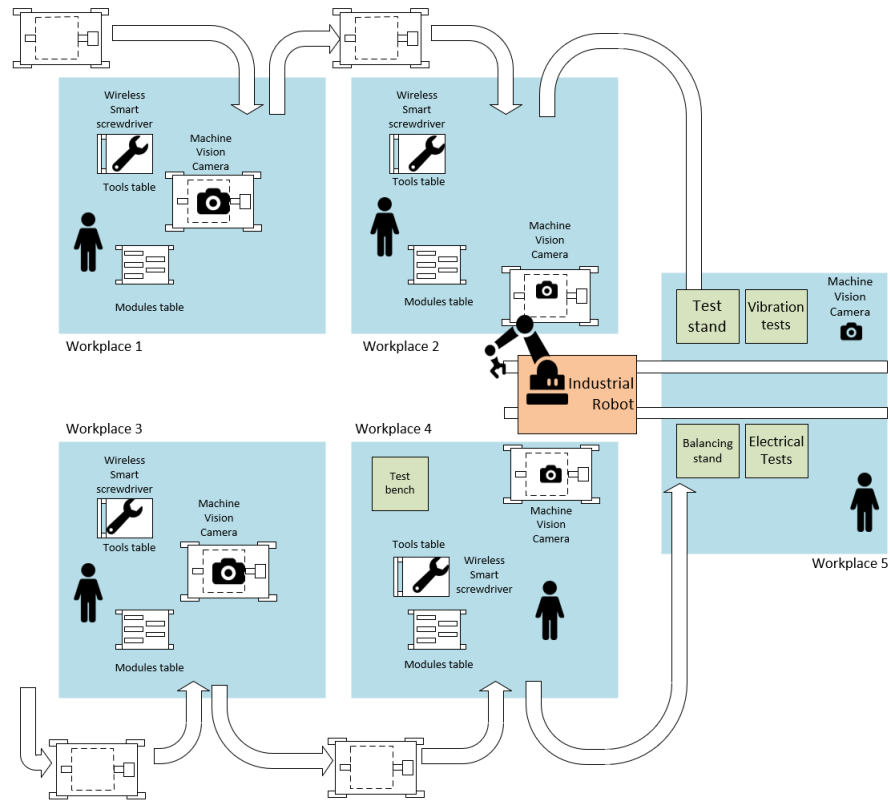

**FIGURE 1.** Scheme of the layout of the work area for the assembly of a small spacecraft (the workplace of the storekeeper is not shown).

Assembly control was automated at the following workplaces in the pilot version of the system:

- Storekeeper's workplace.
- Workplace No. 1 assembly of the main instrument panel.
- Workplace No. 2 subassembly of auxiliary panels.
- Workplace No. 3 of assembly and testing of the solar battery

Since the system considered in this article is used, among other things, for existing workplaces where automation was not supposed to be introduced, there are limitations on the applicability of certain control methods. In particular, equipping workplaces with vision cameras and specialized tools is not possible in all cases.

## TECHNICAL IMPLEMENTATION

The structure of the control system of a separate workplace is shown in Fig. 2.

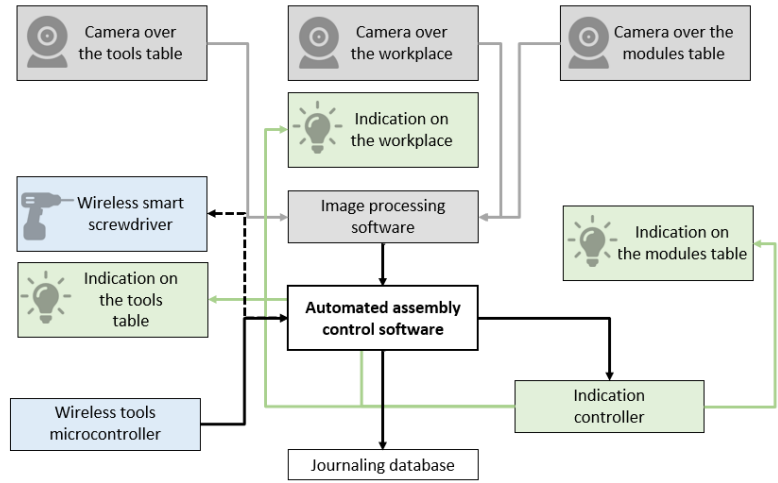

**FIGURE 2.** The structure of the automated workplace control system.

Depending on the workplace, the composition of the equipment may vary, for example, a specialized tool is not required at the warehouseman's workplace. A summary of workplace equipment is shown in Table 1.

**TABLE 1.** The composition of the equipment of workplaces on an experimental assembly line of a small spacecraft.

| Workplace | Machine vision cameras, pcs. | Special tool | Light indication |
|---|---|---|---|
| Warehouse worker | 1 | - | x |
| workplace N 1 | 3 | x | x |
| workplace N 2 | 3 | x | x |
| workplace N 3 | 1 | x | x |

## Automated workplace control system based on machine vision

The most preferable for equipping with a visual control system are workplaces where the assembly object is in a fixed position in the tooling and is not blocked by a worker or other objects. It is also necessary to ensure sufficient illumination of the workplace to avoid shading the area of interest on the camera.

Machine vision cameras are usually located above the workplace at a height of 1..2 m from the assembly table in order not to interfere with the worker. At the same time, to capture the entire table in the frame, it is required to use a camera with a wide angle of view (at least 90 degrees), which entails problems caused by geometry distortion, as well as low resolution per unit area.

The hardware part consists of one or more machine vision cameras, as well as a "smart" tool, which are connected to a PC. The layout used one camera with a resolution of 2 megapixels and a viewing angle of 90 degrees, in the serial system it is planned to use up to three cameras with a resolution of 8 megapixels, which will allow more reliable tracking of the twisting of small assembly units. Also, an 8 megapixel camera was installed at the warehouseman's workplace for reliable recognition of marks on assembly units.

The control system at the workplace performs the following functions:

- Recognizes assembly elements.
- Checks the correct installation of elements.
- Controls tightening torque and controls special tool.
- Records all operator actions in a log

The software provides object recognition using the OpenCV libraries [5], the user interface is implemented using the Qt libraries [6]. To find assembly nodes, the pattern recognition method is used, which was chosen due to low requirements for computing resources [7]. Fig. 3 shows the interface of the workplace control software during assembly.

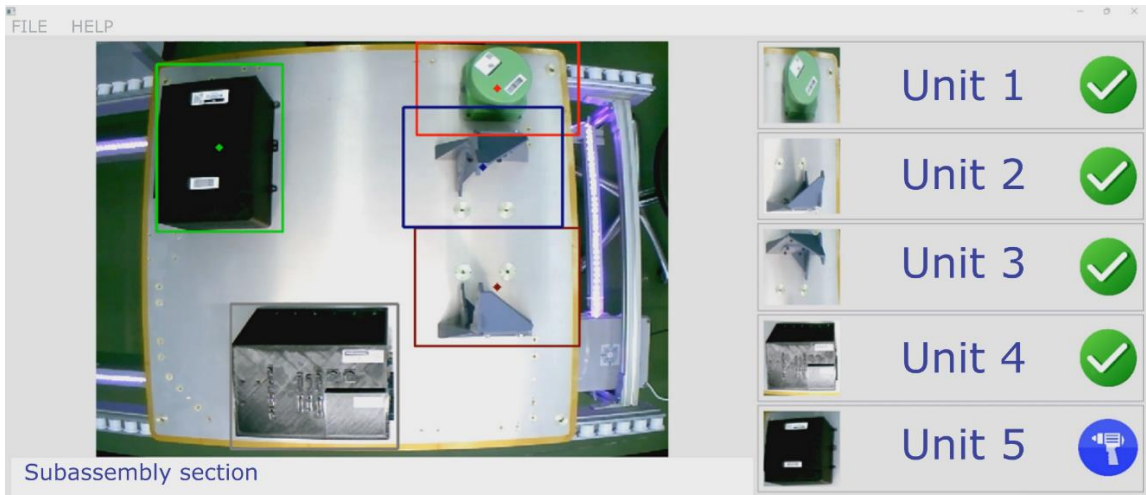

**FIGURE 3.** Workstation control software interface N 1 during assembly.

The software allows the worker to visually control the correct installation of individual elements on the spacecraft panel. In case of an incorrect setting, a warning and a light signal are displayed.

All technological transitions are set in the form of a Python script, where you can specify individual operations, such as installing an element, tightening screws, etc. If it is necessary to tighten the screws, a signal is transmitted to a special tool via a wireless communication channel, which sets the tool for tightening with a certain torque.

## Automated tightening control of screw connections

In any assembly operations where threaded connections are used, a special tool is used to tighten the connection to the required torque. However, most often it is not torque limitation that is used, but only an indication that reports the achievement of the required torque by light indication or mechanically. There are methods for controlling tightening torques based on experimentally obtained characteristics, which require additional control measurement after tightening, which complicates the control process by an order of magnitude [4].

In critical assemblies, where soft materials such as aluminum are used, even when using a tool with an indication, it is possible to make a mistake and overtighten the connection. To eliminate the possibility of this error, the authors proposed and developed a specialized tightening tool with electronic torque limitation and tightening torque control. The use of such a tool makes it possible both to prevent the joint from being tightened and to fix the value of the achieved torque, which can later be entered into the electronic passport of the product or the technological report.

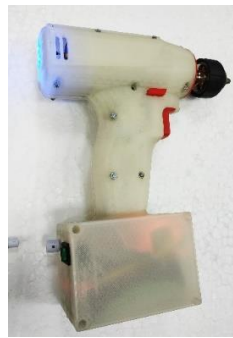 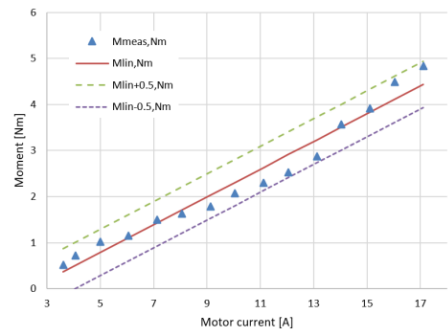

(a) (b)

**FIGURE 4.** (a) Cordless tool with torque measurement and limiting capability (b) Motor output torque versus current. Mmeas – measured values, Mlin – approximation, Mlin+0.5 – approximation +0.5 Nm, Mlin-0.5 – approximation -0.5 Nm.

Figure 4 gives the conversion factor of the motor current to torque, the proportionality factor is equal to

$$K = 0.3 \, Nm/A \tag{1}$$

The graph also shows that with a linear approximation, the deviation of the actual torque is less than ±0.5 Nm, which makes it possible to normalize the torque of most threaded connections with sufficient accuracy. The obtained dependence corresponds to the results obtained in other sources [8,9].

The hardware implementation of the tool allows you to limit the current at a given level, in which case the tool operates in the torque limit mode, or set the tool actuation level, after which the twisting is automatically interrupted. The tool has a wireless communication interface, which gives the worker more freedom of action when accessing hard-to-reach places.

After the successful completion of each of the assembly stages, a log of completed work is generated, which is saved and sent to the database server for entry into the digital passport of the product and analysis of possible erroneous actions or delays of the worker.

An additional advantage that automation of assembly control provides is the ability to obtain the exact state of the product at each stage of assembly, including video recording and measurement of parameters. Due to the large amount of data, it is required to limit the duration of storage of video materials from workplaces. It is advisable to store video information for no more than six months to detect problems in the production technology, and then limit yourself to individual frames at different stages of assembly and compressed copies of the video stream.

## CONCLUSION

The system proposed makes it possible to provide almost any small-scale or mass production with the possibility of automated control of any assembly operations, regardless of their degree of automation. The system can be implemented both in existing industries and in newly created ones. One of the features of the proposed system is a distributed architecture, in which most of the operations are provided by a locally installed computer at the workplace, and a connection to the database is required only for logging and updating the technological process. Advanced features, such as light indication, display of hints to the operator, make it possible to simplify the process of assembling complex products.

The developed method for limiting the tightening torque using a special tool allows using one sensor to provide both hardware limitation and direct measurement of the tightening torque during operation, which simplifies and reduces the cost of the system.

Further development of the system is the addition of customization options for workplaces, including machine vision training, marking of work areas, adding technological transitions and nodes, to enable the system to be configured by its own technological departments and services.

## ACKNOWLEDGMENTS

This work was supported by the Ministry of Science and Higher Education of the Russian Federation (State Contract No. FEFE-2020-0017).